\pdfoutput=1

\documentclass[11pt]{article}

\usepackage[]{ACL2023}

\usepackage{arydshln}
\usepackage{times}
\usepackage{latexsym}
\usepackage[utf8]{inputenc}
\usepackage[mathletters]{ucs}
\usepackage{hyperref}
\usepackage{url}
\usepackage{graphicx}
\usepackage{tabularx}
\usepackage{wrapfig}
\usepackage{booktabs}
\usepackage{multirow}
\usepackage{xspace}
\usepackage{tablefootnote}
\usepackage{caption}
\usepackage{subcaption}
\usepackage{algorithm}
\usepackage{algpseudocode}
\usepackage{euflag}
\usepackage{wrapfig}
\usepackage{pifont}
\newcommand{\cmark}{\ding{51}}%
\newcommand{\xmark}{\ding{55}}%
\usepackage{inconsolata}
\usepackage{fdsymbol}
\usepackage{soul}

\usepackage{siunitx}
\usepackage{colortbl}
\colorlet{tableheadcolor}{gray!75}
\colorlet{tablerowcolor}{gray!10}
\newcommand{\rowcol}{\rowcolor{tablerowcolor}} %

\newcommand{\circa}{{\raise.17ex\hbox{$\scriptstyle\sim$}}}

\newcommand\method{SSVP-SLT\xspace}

\title{Towards Privacy-Aware Sign Language Translation at Scale}

\author{\textbf{Phillip Rust$^{\varheartsuit,\dagger}$ \hspace{20pt} Bowen Shi$^{\diamondsuit}$ \hspace{20pt} Skyler Wang$^{\diamondsuit\clubsuit}$}\\
\textbf{Necati Cihan Camg\"oz$^{\vardiamondsuit}$ \hspace{20pt} Jean Maillard$^{\diamondsuit,\ddagger}$}\vspace{0.3cm}\\
\text{$^{\varheartsuit}$ University of Copenhagen \hspace{10pt} $^{\diamondsuit}$ FAIR at Meta \hspace{10pt} $^{\vardiamondsuit}$ Meta \hspace{10pt}  $^{\clubsuit}$ UC Berkeley}\\
\text{$^\dagger$ \texttt{p.rust@di.ku.dk} \hspace{10pt} $^\ddagger$ \texttt{jeanm@meta.com}}
}

\begin{document}
\maketitle
\begin{abstract}
A major impediment to the advancement of sign language translation (SLT) is data scarcity. Much of the sign language data currently available on the web cannot be used for training supervised models due to the lack of aligned captions. Furthermore, scaling SLT using large-scale web-scraped datasets bears privacy risks due to the presence of biometric information, which the responsible development of SLT technologies should account for.
In this work, we propose a two-stage framework for privacy-aware SLT at scale that addresses both of these issues.
We introduce \method, which leverages self-supervised video pretraining on anonymized and unannotated videos, followed by supervised SLT finetuning on a curated parallel dataset. \method achieves state-of-the-art finetuned and zero-shot gloss-free SLT performance on the How2Sign dataset, outperforming the strongest respective baselines by over 3 BLEU-4. Based on controlled experiments, we further discuss the advantages and limitations of self-supervised pretraining and anonymization via facial obfuscation for SLT.

\vspace{.3em}
\includegraphics[width=1.25em,height=1.25em]{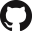}{\hspace{.75em}\parbox{\dimexpr\linewidth-2\fboxsep-2\fboxrule}{\vspace{-5pt} \href{https://github.com/facebookresearch/ssvp_slt}{\texttt{facebookresearch/ssvp\_slt}}}}

\end{abstract}

\section{Introduction}

Used by millions worldwide, sign languages play a crucial role in facilitating communication for many d/Deaf and hard-of-hearing individuals.
Visual in nature, these languages make use of the co-articulated features of hands (i.e., finger positioning, shape, movement, palm orientation, etc.), body postures, gaze, mouth gestures, mouthings and facial expressions to convey meaning \citep{stokoe1980sign}. Globally, there are more than \num{300} sign languages, each with their own grammar and vocabulary.\footnote{\url{https://www.un.org/en/observances/sign-languages-day}} American Sign Language (ASL) alone is estimated to have more than half a million native users, ranking it among the most commonly used languages in the United States \citep{mitchell-etal-2022-how}.

\begin{table}[t]
\centering
\fontsize{10}{12}\selectfont
\sisetup{table-format = 3.2}
\resizebox{0.4\textwidth}{!}{%
\begin{tabular}{@{}llll@{}}
\toprule
\multicolumn{2}{l}{\textsc{Stage}} & \textsc{I.} & \textsc{II.} \\ \midrule
\multicolumn{4}{l}{\textsc{Data}} \\
& \textsc{Scale} & Large & Smaller \\
& \textsc{Source} & Web-mined & Hand-curated \\
& \textsc{Annotated} & \xmark & \cmark \\
& \textsc{Anonymized} & \cmark & \begin{tabular}[c]{@{}l@{}}\cmark / \xmark \textsuperscript{  (with explicit consent)}\end{tabular} \\ \midrule
\multicolumn{2}{l}{\textsc{Training}} & Self-supervised & Supervised \\ \midrule
\multicolumn{2}{l}{\textsc{Output}} & \begin{tabular}[c]{@{}l@{}}Pretrained model\end{tabular} & Translations \\ \bottomrule
\end{tabular}%
}
\caption{Our proposed generic, scalable and privacy-aware SLT framework. We make no assumptions about model architecture and anonymization method.}
\label{tab:framework}
\end{table}

Despite the prevalence of sign languages, they are still under-served by translation technology. Besides under-investment \citep{yin-etal-2021-including} and the inherent difficulty of SLT,\footnote{Results of the WMT~2023 SLT task evince this difficulty; the best system only achieved \circa 1~BLEU \citep{muller-etal-2023-finding}.} another key explanation for this imbalance is the lack of sufficiently large, clean, and labeled parallel corpora. Current state-of-the-art SLT systems require detailed and time aligned annotations \citep{zhou-etal-2023-iccv, uthus-etal-2023-youtube}, which is not scalable, as annotating sign language data is a labour intensive task and can only be done by proficient signers.

We argue that a promising solution to SLT's data scarcity is to utilize publicly available \emph{unannotated} sign language data.\footnote{For example, \citet{uthus-etal-2023-youtube} filtered their Youtube-ASL dataset from \circa $88$K to $11$K videos based largely on the availability and quality of English captions.}
In other domains of computer vision and NLP, a common practice is to pretrain on large-scale unannotated web datasets and later finetune on curated, task-specific datasets \citep{devlin-etal-2019, radford-etal-2018-improving, radford-etal-2019-language, raffel-etal-2020-t5, he-etal-2022-mae}. This practice is largely unexplored in the SLT domain and comes with additional challenges. In particular, moving to large-scale sign language processing makes it increasingly difficult to control the composition of the training data. Because sign language videos typically feature faces and upper bodies and thus are biometrically identifying, such research may exacerbate privacy risks. Hence, developing sign language technologies responsibly requires us to account for these risks and explore techniques to protect privacy.

In this work, we study the effectiveness of self-supervised video pretraining for SLT, under consideration of the aforementioned privacy risks. We first propose a \textit{generic, scalable and privacy-aware} two-stage framework for SLT, summarized in Table~\ref{tab:framework}. We introduce \textbf{\textit{\method}} (\textbf{S}elf-\textbf{S}upervised \textbf{V}ideo \textbf{P}retraining for \textbf{S}ign \textbf{L}anguage \textbf{T}ranslation), an implementation of this framework consisting of two or optionally three stages: pretraining a continuous sign language encoder via masked autoencoding \citep[MAE; ][]{he-etal-2022-mae} on anonymized video, then optionally bridging the modality gap via CLIP-style video-text pretraining \citep{radford-etal-2021-clip}, and finally training an SLT system via supervised finetuning using extracted features from the pretrained model. Our best performing models achieve \num{15.5} BLEU finetuned and \num{7.1} BLEU zero-shot on the How2Sign dataset \citep{duarte-etal-2021-how2sign}, surpassing SOTA in both settings by over \num{3}~BLEU while using data anonymized via facial obfuscation. We also introduce a new ASL-to-English SLT benchmark dataset, \emph{DailyMoth-70h}, consisting of \circa 70h of continuous signing in native ASL. We then evaluate the downstream performance impact and discuss the benefits and limitations of facial blurring to achieve anonymization. Through controlled ablation studies of \method, we identify what factors contribute to a strong pretraining and finetuning recipe. We conclude by discussing opportunities and challenges of self-supervised pretraining for sign language processing.

\begin{figure*}[ht!]
\centering
\includegraphics[width=\textwidth]{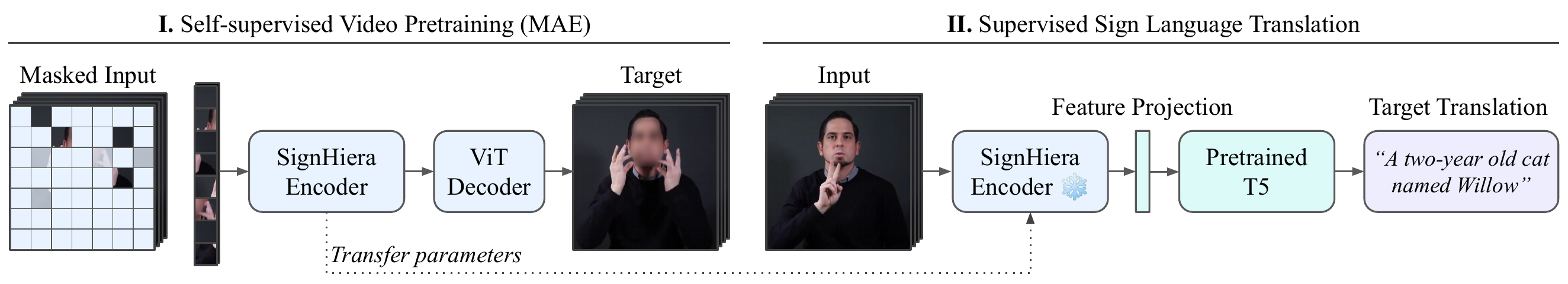}
\caption{Overview of our two-stage \method method. The first stage consists of training a SignHiera encoder via masked autoencoding (MAE) on \emph{blurred} video frames.
In the second stage, a pretrained T5 model is finetuned for SLT while the pretrained SignHiera is kept frozen (\includegraphics[width=0.9em]{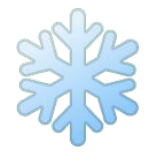}).
The input video in the second stage \emph{can be unblurred}.
}
\label{fig:method_figure}
\end{figure*}

\section{Background and Related Work}

\paragraph{Gloss-free SLT}
\label{sec:gloss_free_translation}

Glosses are a way of representing individual signs into a written form. Being monotonically aligned to signs, they can be a useful medium between sign and spoken languages. Most SLT approaches to date rely on them \citep{chen-etal-2022-simple, chen-etal-2022-twostream, zhang-etal-2023-sltunet}. The task of predicting glosses from continuous signing is typically performed via gloss supervision either jointly or in a cascaded approach with supervised SLT finetuning \citep{camgoz-etal-2018-neural, camgoz-etal-2020-sign}.

However, glosses are also considered an incomplete and inaccurate representation of sign language \citep{camgoz-etal-2020-sign, mueller-etal-2023-considerations}. Furthermore gloss annotation is a labour intensive task. Due to these constraints, there is a growing body of research on gloss-free SLT. Most such approaches incorporate techniques aimed at reducing the modality gap, such as training the visual encoder via sign spotting \citep{tarres-etal-2023-instructional, shi-etal-2022-open},\footnote{While \citet{tarres-etal-2023-instructional} do not explicitly mention the use of sign spotting, they rely on features extracted from an I3D model \citep{carreira-etal-2017-i3d} by \citet{duarte-etal-2022-sign}, who used an iterative sign spotting technique.} adding inductive bias in the attention mechanism \citep{yin-etal-2023-gloss}, using conceptual anchor words \citep{lin-etal-2023-gloss}, or performing visual-language pretraining \citep{zhou-etal-2023-iccv}. \citet{uthus-etal-2023-youtube} also benefit from a pretrained text model \citep[T5; ][]{raffel-etal-2020-t5}. Similar to \citet{zhou-etal-2023-iccv}, we also leverage language supervision to reduce the modality gap, albeit in conjunction with self-supervised video pretraining.

\paragraph{Sign Language Video Anonymization}
\label{sec:sign_anonymization}

Sign language videos typically feature signers' faces, which convey essential linguistic information. However, in virtual domains, particularly in spaces involving the discussion of sensitive topics, exposing one's face (and identity) may lead to various forms of personal risks. Such exposures could even lead to harm associated with professional or insurance discrimination.
For these reasons, the d/Deaf and hard-of-hearing community has long expressed interest in anonymization and privacy protection techniques for sign language content \citep{lee-etal-2021-anonymization}, and sign language video anonymization has, in recent years, become an active area of research \citep{isard-2020-approaches, xia-etal-2023-diffslva}.

For general-domain images and videos, simple anonymization techniques such as facial obfuscation via overlaying or blurring are widely used and accepted \citep{frome-etal-2009-streetview,yang-etal-2022-face}. In the sign language domain, such techniques may be inadequate due to the loss of information in the facial region \citep{lee-etal-2021-anonymization}. More specifically, in signed language, non-manual features such as mouthing, eyebrow and head movements are used extensively to enrich grammar. Certain signs with similar manual features are only disambiguated through mouth morphemes. Moreover, facial expressions are often used to indicate emphasis, negation, a question, etc. \citep{baker-shenk-1985-facial, valli-lucas-2000-linguistics, neidle-etal-2000-syntax}.

Recent work has focused on anonymizing sign language content via avatars \citep{tze-etal-2022-cartoonized, lee-etal-2021-anonymization} and transferred or synthetic human appearances \citep{saunders-etal-2021-anonysign, xia-etal-2022-sign, xia-etal-2023-diffslva}. While promising, these approaches are nascent and we are unaware of studies that determine to what extent models can learn to recover or disambiguate obfuscated linguistic information from context. That being said, human studies suggest that signers struggle to comprehend content anonymized in such a way \citep{lee-etal-2021-anonymization}.

Lacking an obvious alternative, in this work we return to the relatively straightforward technique of facial blurring. Despite its limitations, we demonstrate that blurring can raise privacy protection with little performance degradation.\footnote{In Appendix~\ref{app:pose_vs_video} we discuss issues with pose landmarks, often promoted as a privacy-preserving alternative to video.} This, we argue, can facilitate large-scale video anonymization when applied to publicly available sign language data.

\paragraph{Masked Autoencoding for Video and Beyond}
Following its success as a self-supervised learning paradigm in the image domain \citep{he-etal-2022-mae}, MAE has been widely applied in other areas, including audio \citep{huang-etal-2022-mae}, language \citep{rust-etal-2023-pixel}, and video \citep{feichtenhofer-etal-2022-maest, tong-etal-2022-videomae, wang-etal-2023-videomaev2}.
Considering that MAEs have been shown to be capable of acquiring both language and basic video understanding from pixels alone, it is conceivable that high-quality sign language representations can be learned directly from RGB video data via MAE, given enough data. Recently, \citet{sandoval-castaneda-etal-2023-video} explored MAE among other self-supervised learning techniques in the context of isolated sign language recognition (ISLR) and found it to be highly useful. MAE has, however, not yet been successfully applied to SLT. In SLT, videos are much longer, and learning high-quality representations requires models to capture long-range spatiotemporal dependencies. Our usage of MAE, or self-supervised pretraining in general, therefore stands in contrast to recent SLT methods, gloss-based and gloss-free methods alike, which instead fully rely on supervised training that requires annotated captions \citep{zhou-etal-2023-iccv, tarres-etal-2023-instructional, lin-etal-2023-gloss, uthus-etal-2023-youtube}.

\section{Generic Framework}

We first outline a generic, scalable and privacy-aware two-stage transfer learning framework for SLT (See Table~\ref{tab:framework}).

In \textbf{Stage I}, we train a model, with the goal to learn high-quality continuous sign language
representations, via self-supervised learning. The data used at this stage is \emph{always anonymized}. We make no assumptions on how the data may be anonymized, i.e. face blurring as discussed in \S\ref{sec:sign_anonymization}, or more sophisticated methods such as using synthetic appearances.

In \textbf{Stage II}, we finetune the model from the first stage in a supervised manner using a smaller and hand-curated parallel dataset. Ideally, the finetuning dataset, being manageable in size, can be unanonymized after gaining explicit consent from signers in the data to minimize information loss.

\section{Method}

The base implementation of our framework is designed as a two-step approach, termed \textbf{\method}. We provide a high-level overview in Fig.~\ref{fig:method_figure}.

\paragraph{Self-Supervised Video Pretraining (MAE)}
We first aimed to pretrain a capable sign language encoder on video data alone---no gloss, pseudo-gloss, or spoken-language text annotations---allowing us to leverage large amounts of unannotated sign language video in the future, alleviating the data scarcity issue in SLT training.

Our sign video encoder, \emph{SignHiera}, builds on Hiera \citep{ryali-etal-2023-hiera}, a vision transformer that combines a hierarchical architecture, shown to be crucial for learning phonetically meaningful sign representations \citep{sandoval-castaneda-etal-2023-video} with masked autoencoding (MAE), a widely used self-supervised learning paradigm \citep{he-etal-2022-mae}. Its hierarchical architecture also makes Hiera more efficient to train than other MAE-based video transformers such as VideoMAE \citep{tong-etal-2022-videomae, wang-etal-2023-videomaev2} or MAE-ST \citep{feichtenhofer-etal-2022-maest}.

Hiera embeds a video into a sequence of spatio-temporal tokens. A large percentage of tokens is randomly masked, while the rest is passed through a hierarchical transformer stack with several pooling operations. The decoder receives fused multi-scale features extracted before each pooling operation and processes them via a lightweight transformer stack. A final linear projection yields pixel logits. The loss is computed as the normalized mean squared error between the original and predicted pixel values of the masked tokens.

SignHiera is initialized from the original Hiera-Base-\num{16}$\times$\num{224} checkpoint pretrained on Kinetics-\num{400} \citep{kay-etal-2017-kinetics}. In order to capture longer-range spatio-temporal dependencies in signed utterances, we increased the clip length from \num{16} to \num{128} video frames, leading to an $8\times$ sequence length, and accordingly resized and reinitialized the position embeddings. We further added attention masking to accommodate shorter videos with temporal padding and added a third Q-pooling operation after the last encoder stage to save GPU memory. We trained with a masking ratio of \num{0.9}.

\paragraph{Supervised SLT Finetuning}

The translation network is an encoder-decoder transformer model \citep{vaswani-etal-2017}. Our default configuration uses a pretrained T5-v1.1 \citep{raffel-etal-2020-t5, shazeer-2020-glu}, following \citet{uthus-etal-2023-youtube}. We also experimented with BART \citep{lewis-etal-2020-bart} and \citet{tarres-etal-2023-instructional}'s setup, training a \circa \num{10}M parameter transformer from scratch.

The only difference from a text transformer is that our translation network takes in video features extracted from the pretrained SignHiera. We used SignHiera's final intermediate representations, which are of size $\mathbb{R}^{B \times \frac{T}{2} \times \frac{H}{32} \times \frac{W}{32} \times D}$, where $B$ is the batch size, $T$=$128$, $H$=$W$=$224$ is the input video size, and $D$=$768$ is the feature size. We mean-pooled over the spatial dimensions to obtain feature vectors of size $\mathbb{R}^{B \times \frac{T}{2} \times D}$. Videos shorter than \num{128} frames were padded for encoding, and the padding was then removed from the extracted features. For longer videos, we used a sliding window with stride $\frac{T}{2}$ and concatenated the resulting features. A linear projection $W_{\text{proj}} \in \mathbb{R}^{D \times D'}$
mapped the extracted features to a sequence of size $\mathbb{R}^{B \times S \times D'}$, with $S$ being the sequence length of the extracted features and $D'$ the transformer's hidden size. This sequence was processed by the transformer as usual \citep{vaswani-etal-2017}.

\begin{figure}[ht!]
\centering
\includegraphics[width=0.48\textwidth]{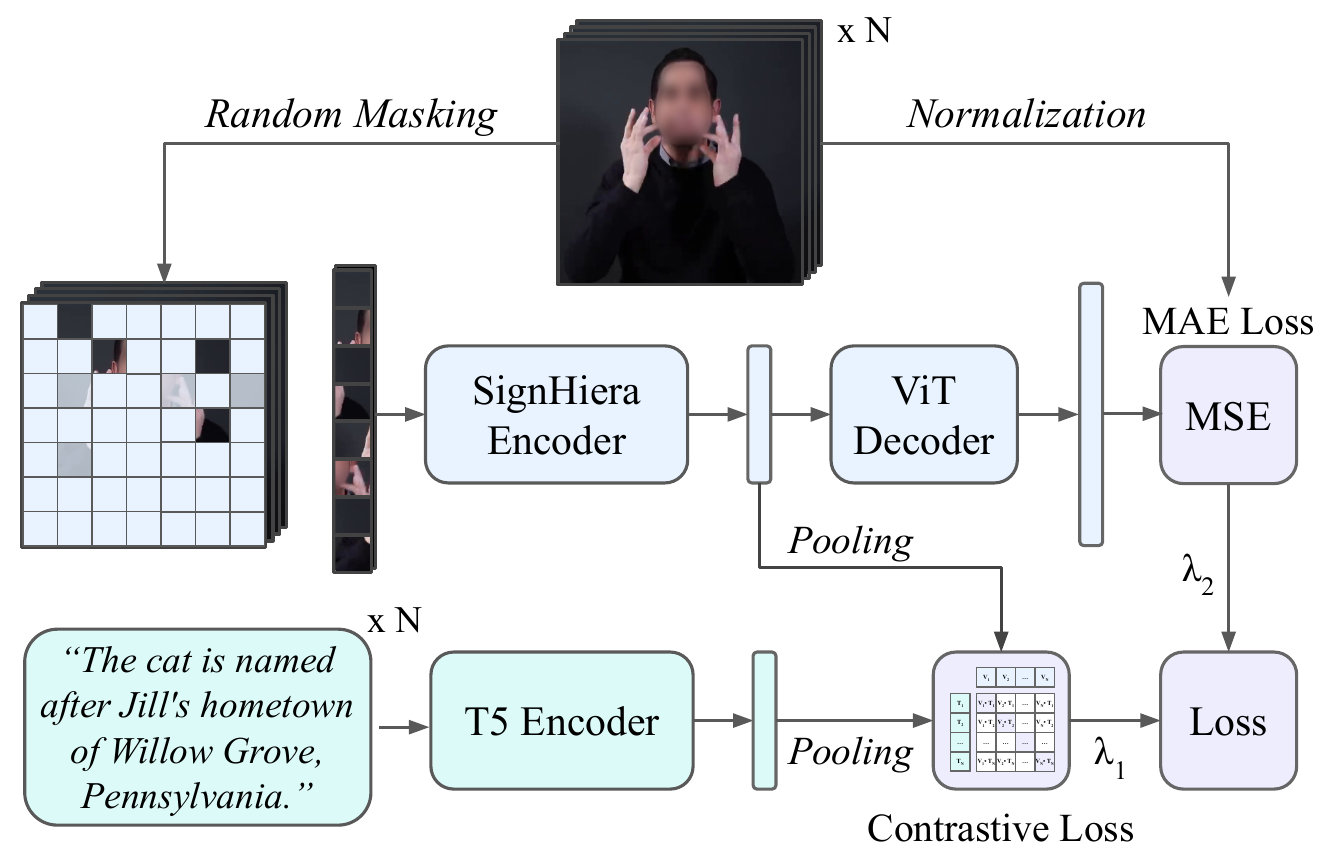}
\caption{Overview of our LSP extension.}
\label{fig:method_figure_language_supervision}
\end{figure}

\paragraph{Adding Language-supervised Pretraining}
We also experimented with extending \method with a language-supervised pretraining (LSP) step to bridge the modality gap between the input videos and text translations. Bridging this gap may improve gloss-free SLT performance, as discussed in \S\ref{sec:gloss_free_translation}.
In what follows, we refer to \method with an additional LSP stage as \textbf{\method-LSP}. The LSP stage fits \emph{in between} the self-supervised MAE pretraining and the supervised SLT finetuning stage. Since language supervision requires annotations, the LSP stage should be considered a part of stage II in our generic framework.

Our LSP approach is illustrated in Figure~\ref{fig:method_figure_language_supervision}.
We first initialized a CLIP model \citep{radford-etal-2021-clip} with our MAE-pretrained SignHiera encoder as the vision tower and a pretrained T5-v1.1 encoder as the text tower. We then jointly trained the CLIP model via contrastive video-text pretraining and the SignHiera model via MAE. The goal is to help the SignHiera encoder, which is involved in both tasks, learn strong continuous sign representations grounded in the target modality (text). The videos were masked with a 90\% ratio even for computing the contrastive loss, which is similar to FLIP \citep{li-etal-2023-flip}, and enabled end-to-end training by drastically reducing the memory footprint. The two losses (MAE and contrastive) were balanced via GradNorm \citep{chen-etal-2018-gradnorm}, which helped stabilize training, compared to using fixed loss weights.\footnote{In contrast to \citet{zhou-etal-2023-iccv}, we did not jointly train the text decoder as doing so did not improve performance in preliminary experiments and led to training instabilities.}

\section{Experimental Setup}

\subsection{Datasets}
\paragraph{Youtube-ASL (YT)} We used Youtube-ASL \citep{uthus-etal-2023-youtube}, the largest available ASL training dataset with \circa 1000h of in-the-wild video and over 2500 signers, both during pretraining and supervised finetuning.\footnote{For the lack of a readily available larger, unannotated dataset, Youtube-ASL fits both dimensions of our framework: the large, publicly-available, unannotated dataset and the curated, parallel dataset.} We used a version in which all signers' faces are blurred for privacy.

\paragraph{How2Sign (H2S)} We also used How2Sign  \citep{duarte-etal-2021-how2sign}, an ASL dataset with \circa 80h of video and nine different signers in a green screen studio setting, for pretraining, finetuning, and downstream evaluation of our SLT models. Again, we used a version with blurred faces only.

\paragraph{DailyMoth-70h} To isolate the impact of face blurring during pretraining and finetuning on SLT performance, we relied on a new dataset, which we name \emph{DailyMoth-70h}. This dataset contains over 70h of video of a single signer from the ASL news page TheDailyMoth and was obtained through a license agreement with TheDailyMoth's host.\footnote{\url{https://www.dailymoth.com/}} We used both unblurred and blurred dataset versions and report dataset statistics in Appendix~\ref{app:dailymoth_data}.

\begin{table*}[ht!]
\centering
\fontsize{10}{12}\selectfont
\sisetup{table-format = 3.2}
\resizebox{\textwidth}{!}{%
\begin{tabular}{@{}lclcccccc@{}}
\toprule
\textsc{Method} & \textsc{Blur} & \textsc{FT Data} & \multicolumn{1}{l}{\textsc{BLEU-1}} & \textsc{BLEU-2} & \textsc{BLEU-3} & \textsc{BLEU} & \textsc{ROUGE-L} & \multicolumn{1}{l}{\textsc{BLEURT}} \\
\midrule
\addlinespace[0.3em]
\rowcol\multicolumn{9}{c}{\textbf{Baselines}} \\
\addlinespace[0.2em]
\citet{lin-etal-2023-gloss} & \xmark & H2S & \num{14.9} & \num{7.3} & \num{3.9} & \num{2.2} & \num{12.6} & \num{31.7} \\
\citet{tarres-etal-2023-instructional} & \xmark & H2S & \num{34.0} & \num{19.3} & \num{12.2} & \num{8.0} & --- & --- \\
\addlinespace[0.3em]
\cdashline{0-8}
\addlinespace[0.3em]
\multirow{4}{*}{\citet{uthus-etal-2023-youtube}} & \multirow{4}{*}{\xmark} & H2S & \num{15.0} & \num{5.1} & \num{2.3} & \num{1.2} & --- & \num{30.0} \\
&  & YT & \num{20.9} & \num{10.4} & \num{6.1} & \num{4.0} & --- & \num{35.0} \\
&  & YT $+$ H2S & \num{36.3} & \num{23.0} & \num{16.1} & \num{11.9} & --- & \num{44.8} \\
&  & YT $\rightarrow$ H2S & \num{37.8} & \num{24.1} & \num{16.9} & \num{12.4} & --- & \num{46.6} \\
\addlinespace[0.3em]
\rowcol\multicolumn{9}{c}{\textbf{Ours}} \\
\addlinespace[0.2em]
\method $_{800}^\text{H2S}$ & \cmark & H2S & \num{30.2} & \num{16.7} & \num{10.5} & \num{7.0} & \num{25.7} & \num{39.3} \\
\addlinespace[0.3em]
\cdashline{0-8}
\addlinespace[0.3em]
\multirow{4}{*}{\method $_{800}^\text{YT}$} & \multirow{4}{*}{\cmark} & H2S & \num{38.1} & \num{23.7} & \num{16.3} & \num{11.7} & \num{33.8} & \num{44.2} \\
&  & YT & \num{29.2} & \num{16.6} & \num{10.7} & \num{7.1} & \num{28.3} & \num{41.8} \\
&  & YT $+$ H2S & \num{41.6} & \num{27.2} & \num{19.3} & \num{14.3} & \num{36.8} & \num{48.6} \\
&  & YT $\rightarrow$ H2S & \num{41.9} & \num{27.7} & \num{19.8} & \num{14.7} & \num{37.8} & \num{49.3} \\
\addlinespace[0.3em]
\cdashline{0-8}
\addlinespace[0.3em]
\multirow{4}{*}{\method $_{800}^{\text{YT} + \text{H2S}}$} & \multirow{4}{*}{\cmark} & H2S & \num{38.9} & \num{24.1} & \num{16.4} & \num{11.6} & \num{34.0} & \num{44.5} \\
&  & YT & \num{29.1} & \num{17.0} & \num{11.1} & \num{7.5} & \num{28.6} & \num{41.6} \\
&  & YT $+$ H2S & \num{41.8} & \num{27.4} & \num{19.5} & \num{14.3} & \num{36.9} & \num{48.9} \\
&  & YT $\rightarrow$ H2S & \num{41.8} & \num{27.4} & \num{19.6} & \num{14.6} & \num{37.7} & \num{49.0} \\
\addlinespace[0.3em]
\cdashline{0-8}
\addlinespace[0.3em]
\method-\textsc{LSP} $_{600 \rightarrow 200}^{\text{YT} + \text{H2S}}$ & \cmark & YT $+$ H2S & $\mathbf{43.2}$ & $\mathbf{28.8}$ & $\mathbf{20.8}$ & $\mathbf{15.5}$ & $\mathbf{38.4}$ & $\mathbf{49.6}$ \\
\bottomrule
\end{tabular}%
}
\caption{How2Sign test performance of \method in different pretraining configurations compared to baselines. The \textsc{blur} column denotes whether faces in the train and eval data are blurred. \textsc{ft data} indicates the finetuning configuration; respectively, YT+H2S and YT$\rightarrow$H2S refer to training on the two datasets jointly or consecutively.}
\label{tab:main_results}
\end{table*}

\subsection{Training and Evaluation Protocols}
We briefly describe our training and evaluation protocols. The full configurations for pretraining and SLT training are listed in Appendix~\ref{app:training_details}.

\paragraph{Face blurring} We used an internal face blurring software and ensured its reliability on the YT and H2S datasets via a combination of automatic and manual verification techniques. Example frames sampled from two blurred videos from the DailyMoth-70h data are shown in Appendix Figure~\ref{fig:example_blurred_frames}.

\paragraph{MAE pretraining} We largely follow the Hiera pretraining recipe from \citet{ryali-etal-2023-hiera}. In our default configuration, we trained for \num{800} effective epochs (accounting for repeated sampling as in \citet{feichtenhofer-etal-2022-maest}) with the AdamW optimizer \citep{kingma-ba-2015-adam, loshchilov-hutter-2018-adamw}. The learning rate was set to \num{8e-4} with linear warmup over \num{120} epochs and cosine decay to \num{1e-5}. The batch size was $2\times128$, with \num{2} being the repeated sampling factor. Similar to \citet{zhou-etal-2023-iccv}, we employed video data augmentation via random cropping, horizontal flipping, and RandAug \citep{cubak-etal-2020-randaug}. We used a $128 \times 2$ temporal sampling strategy, i.e., sampling \num{128} frames with a stride of $2$, which fully accommodates \circa 85--95\% of videos in the data.

\paragraph{SLT finetuning}
When finetuning only on How2Sign or DailyMoth-70h, we closely followed the setup of \citet{tarres-etal-2023-instructional}, training a \circa \num{10}M parameters transformer from scratch; see Appendix~\ref{app:training_details} for more details.
For How2Sign, we reused their lowercase unigram tokenizer (vocab size \num{7}K). For DailyMoth-\num{70}h, we trained a cased tokenizer (unigram, \num{7}K vocab size), which we found to work better due to the large proportion of named entities in the data.

When finetuning on Youtube-ASL, as we needed a model with more capacity we relied on a pretrained T5-v1.1 with default tokenizer, following \citet{uthus-etal-2023-youtube}. We trained for up to \num{100} epochs with early stopping, batch size \num{512} and AdamW with peak rate \num{5e-4}, linear warmup over two epochs and cosine decay to \num{1e-7}. We used dropout of \num{0.3} and \num{0.2} label smoothing in the cross-entropy loss. We did not use video data augmentation unless specified otherwise.

\paragraph{Language-supervised pretraining}
We performed \num{200} epochs of LSP with a batch size of $512$ on top of \num{600} epochs of MAE-only pretraining. We did not use repeated sampling, which is incompatible with the contrastive loss. We again used AdamW, warming up to a learning rate of \num{1e-4} over ten epochs, followed by cosine decay to \num{1e-6}. The GradNorm optimizer has a one epoch warmup, a peak learning rate of \num{1e-2}, and decays to \num{1e-4}. Data augmentation and temporal sampling are the same as for MAE pretraining.

\paragraph{Evaluation} We used beam search with \num{5} beams and no length penality. We evaluated every epoch when finetuning on How2Sign or Dailymoth-70h and every \num{500} steps for Youtube-ASL. We kept the checkpoint with the highest validation BLEU-4 and evaluated it on the respective test set.

\paragraph{Notation} Below, we use superscript and subscript to indicate the pretraining dataset and number of epochs, respectively. For instance, \method $_{800}^{\text{YT} + \text{H2S}}$ refers to 800 epochs of MAE pretraining on Youtube-ASL and How2Sign. For \method-LSP, \num{600}$\rightarrow$\num{200} denotes \num{600} epochs of MAE pretraining followed by \num{200} epochs of LSP.

\subsection{Baselines}

\textbf{\citet{lin-etal-2023-gloss}} propose to bridge the visual and text modalities via contrastive anchoring of encoded visual features to embeddings of conceptual words in the target sequence.
\textbf{\citet{tarres-etal-2023-instructional}} is the SOTA on How2Sign without additional SLT data, training a 6+3 layer transformer from scratch on features from an I3D model \citep{carreira-etal-2017-i3d}. The I3D model was first pretrained on Kinetics \citep{carreira-etal-2017-i3d} and BOBSL \citep{albanie-etal-2021-bobsl}, and finetuned on How2Sign for sign language recognition using annotations generated via iterative sign spotting \citep{duarte-etal-2022-sign}.
\textbf{\citet{uthus-etal-2023-youtube}} is the current SOTA on How2Sign, and finetunes a pretrained T5-v1.1-Base model for SLT directly on pose landmarks extracted from YouTube-ASL and How2Sign videos.

\subsection{Evaluation Measures} We report BLEU via SacreBLEU \citep{papineni-etal-2002-bleu, post-etal-2018-sacrebleu}.\footnote{\texttt{nrefs:1|case:mixed|\\eff:no|tok:13a|smooth:exp|version:2.3.1}} We also report ROUGE-L \citep{lin-2004-rouge} and BLEURT \citep{sellam-etal-2020-bleurt} from the BLEURT-20 checkpoint, which has been shown to correlate well with human judgments.

\section{Results and Discussion}

\paragraph{Comparison against the state-of-the-art}
\label{sec:main_results}
We present our main results in Table~\ref{tab:main_results}.
Our best models significantly improve over the previous \num{12.4} BLEU state-of-the-art by \citet{uthus-etal-2023-youtube}. \method achieves \num{14.7} and \num{14.6} BLEU when pretraining on YT and YT+H2S respectively. Our best overall model, utilizing \method-LSP, achieves \num{15.5} BLEU, a \num{3.1} point improvement over the baseline. When pretraining and finetuning on YT only, we also observe a \num{3.1} BLEU improvement (\num{4.0} vs. \num{7.1}) over the previous SOTA in the zero-shot setting. These results demonstrate the overall effectiveness of \method and, more broadly, self-supervised pretraining for SLT.

Pretraining on YT+H2S performs almost the same as training on YT only, with the YT-only models even sometimes performing best. Given the distributional gap between the datasets (in-the-wild YT vs. studio H2S video) and the fact that the YT+H2S models consumed more data, this finding is somewhat surprising. While this may be due in part to randomness in the training dynamics, it could also mean that sufficient finetuning can compensate for not accessing the H2S data at pretraining, presumably because the pretraining set is sufficiently large and diverse. This encouraging result suggests that we can pretrain on large data independently of knowing what our finetuning and inference dataset will be---a crucial requirement for practical SLT applications.

We find that YT data is beneficial both for pretraining and finetuning, which emphasizes the importance of training on large and diverse data and suggests that we can expect further gains from scaling to large public unannotated video.

Finally, we find that bridging the modality gap via language-supervised pretraining yields a \num{1.2} BLEU improvement over its MAE-only counterpart. Given enough annotated data, the technique can be employed independently of self-supervised pretraining at little extra cost.

\begin{table}[ht!]
\centering
\fontsize{10}{12}\selectfont
\sisetup{table-format = 3.2}
\resizebox{0.48\textwidth}{!}{%
\begin{tabular}{@{}ccccc@{}}
\toprule
\multicolumn{2}{c}{\textsc{Blur}} & \multirow{2}{*}{\textsc{BLEU}} & \multirow{2}{*}{\textsc{ROUGE-L}} & \multicolumn{1}{l}{\multirow{2}{*}{\textsc{BLEURT}}} \\
\textsc{Pretrain} & \textsc{SLT} &  &  & \multicolumn{1}{l}{} \\ \midrule

\xmark & \xmark & $\mathbf{28.8}$ & $\mathbf{50.9}$ & $\mathbf{51.7}$ \\
\xmark & \cmark & \num{28.1} & \num{50.6} & \num{51.4} \\
\cmark & \xmark & \num{28.2} & \num{50.3} & \num{51.0} \\
\cmark & \cmark & \num{27.5} & \num{49.6} & \num{50.4} \\ \bottomrule
\end{tabular}%
}
\caption{Performance on \emph{unblurred} test data for \method trained and evaluated on DailyMoth-70h with or without facial blurring during pretraining and SLT.}
\label{tab:dailymoth_blur_results}
\end{table}

\paragraph{What's the effect of blurring?}
We isolate the impact of facial obfuscation via blurring on SLT performance by training \method models on DailyMoth-70h with and without blurring during pretraining and SLT training. We report the results in Table~\ref{tab:dailymoth_blur_results}. As expected, performance is best when not blurring (\num{28.8} BLEU) and worst when blurring at finetuning time (\num{28.1} and \num{27.5} BLEU). Crucially, some performance can be recovered after pretraining on blurred data when performing SLT on unblurred data (\num{28.2} BLEU), validating our proposed framework (see Table~\ref{tab:framework}).\footnote{We hypothesize that even more performance could be recovered if the SignHiera video encoder was unfrozen during SLT training, allowing adaptation to the facial information.} This means we can pretrain in a privacy-aware manner without sacrificing too much performance.

\begin{figure}[ht!]
\centering
\includegraphics[width=0.48\textwidth]{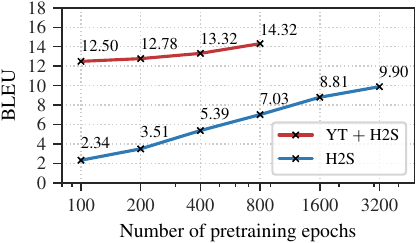}
\caption{How2Sign test BLEU of \method after pretraining on YouTube-ASL and How2Sign or How2Sign only and finetuning on the same data.}
\label{fig:epoch_scaling}
\end{figure}

\paragraph{How long should you pretrain?}
The sign language MAE task is intricate. The model first needs to abstract away surface-level information such as background and signer  appearance. It then needs to implicitly acquire an understanding of ASL, capturing long-range dependencies in the video. It is therefore worth investigating basic scaling properties. In Figure~\ref{fig:epoch_scaling}, we show downstream SLT performance over the course of pretraining. Similar to \citet{ryali-etal-2023-hiera} and \citet{he-etal-2022-mae}, we observe consistent downstream improvements as pretraining progresses, suggesting that the models are not overfitting to the training data even after extensive pretraining. These results underscore the task's effectiveness and indicate that further scaling would likely yield additional gains.

\begin{table}[ht!]
\centering
\fontsize{10}{12}\selectfont
\sisetup{table-format = 3.2}
\resizebox{0.48\textwidth}{!}{%
\begin{tabular}{@{}ccccc@{}}
\toprule
\begin{tabular}[c]{@{}c@{}}\textsc{Clip}\\ \textsc{Size}\end{tabular} & \begin{tabular}[c]{@{}c@{}}\textsc{Pretraining}\\ \textsc{Epochs}\end{tabular} & \textsc{BLEU} & \textsc{ROUGE-L} & \textsc{BLEURT} \\ \midrule
\addlinespace[0.3em]
\rowcol\multicolumn{5}{c}{\textbf{H2S}} \\
\addlinespace[0.2em]
\num{128} & \num{800} & $\mathbf{7.0}$ & $\mathbf{25.7}$ & $\mathbf{39.3}$ \\
\num{128} & \num{100} & \num{2.3} & \num{14.3} & \num{33.6}  \\
\num{16} & \num{800} & \num{5.0} & \num{20.2} & \num{36.4} \\
\addlinespace[0.3em]
\rowcol\multicolumn{5}{c}{\textbf{YT + H2S}} \\
\addlinespace[0.2em]
\num{128} & \num{800} & $\mathbf{14.3}$ & $\mathbf{36.9}$ & $\mathbf{48.9}$ \\
\num{128} & \num{100} & \num{12.5} & \num{34.2} & \num{46.1} \\
\num{16} & \num{800} & \num{10.4} & \num{31.7} & \num{44.3} \\
\bottomrule
\end{tabular}%
}
\caption{How2Sign test performance of \method when pretraining on (YouTube-ASL and) How2Sign  with a clip size of \num{16} versus \num{128} video frames.}
\label{tab:clip_size_ablation}
\end{table}

\paragraph{Is encoding longer clips necessary?}
Increasing clip length is costly due to the poor scaling of global attention, raising the question of whether encoding longer clips is needed. Our results indicate that the answer is \emph{yes}. Table~\ref{tab:clip_size_ablation} compares the performance of the original \num{16}-frame Hiera with our \num{128}-frame SignHiera. While \num{16}-frame Hiera achieves non-trivial performance after \num{800} epochs, it is substantially outperformed by \num{800} epoch SignHiera (\num{7.0} vs \num{5.0} BLEU for H2S and \num{14.3} vs \num{10.4} BLEU for YT+H2S). This may be partially explained by the fact that SignHiera sees up to $8 \times$ as much data every step. However, we also see that \num{800}-epoch \num{16}-frame Hiera lags far behind even a \num{100}-epoch SignHiera (which has seen roughly the same number of tokens) when training on a large dataset (\num{12.5} vs \num{10.4} BLEU in the YT+H2S setting). When training on less data (H2S), \num{16}-frame Hiera is still worse than \num{400}-epoch SignHiera (\num{5.39} vs \num{5.0} BLEU, see Figure~\ref{fig:epoch_scaling}). Overall, this suggests that certain information cannot easily be acquired from shorter clips.

\begin{table}[ht!]
\centering
\fontsize{10}{12}\selectfont
\sisetup{table-format = 3.2}
\resizebox{0.46\textwidth}{!}{%
\begin{tabular}{@{}lccccc@{}}
\toprule
{\textsc{Model}} & {\textsc{Param}} & {\textsc{PT}} & {\textsc{BLEU}} & {\textsc{ROUGE-L}} & {\textsc{BLEURT}} \\ \midrule
\multirow{2}{*}{BART} & \multirow{2}{*}{\num{140}M} & \xmark & \underline{\num{14.0}} & \underline{\num{36.8}} & \underline{\num{48.3}} \\
&  & \cmark & \num{13.5} & \num{36.2} & \num{48.1} \\ \addlinespace[0.3em]
\cdashline{0-5}
\addlinespace[0.3em]
\multirow{2}{*}{T5-v1.1} & \multirow{2}{*}{\num{248}M} & \xmark & \num{11.6} & \num{35.0} & \num{46.1} \\
&  & \cmark & \underline{$\mathbf{14.3}$} & \underline{$\mathbf{36.9}$} & \underline{$\mathbf{48.9}$} \\ \bottomrule
\end{tabular}%
}
\caption{How2Sign test performance of \method~$_{800}^{\text{YT} + \text{H2S}}$ when finetuning BART and T5, initialized randomly (\textsc{PT} = \xmark) or from the pretrained model (\cmark).}
\label{tab:text_model_ablation}
\end{table}

\paragraph{How to choose the text model?}
We investigate how the architecture and initialization of the text transformer affects performance. Table~\ref{tab:text_model_ablation} compares pretrained and randomly initialized BART \citep{lewis-etal-2020-bart}, the English monolingual counterpart to mBART \citep{liu-etal-2020-multilingual-denoising}, which has previously been successfully adapted for German and Chinese sign languages \citep{de-coster-etal-2021-frozen, chen-etal-2022-simple, zhou-etal-2023-iccv}, and T5-v1.1 as used by \citet{uthus-etal-2023-youtube}.

Overall, T5 outperforms BART, possibly due to larger capacity, but the gap is small. While it is worse to finetune a randomly initialized T5 model compared to the pretrained one (corroborating findings by \citet{uthus-etal-2023-youtube}), for BART we find the opposite result. We conclude that whether text pretraining is helpful needs to be evaluated on a case-by-case basis. It may be worth investigating in the future whether an additional pretraining or finetuning step may lead to better adaptation of the text model to sign language.

\begin{table}[ht!]
\centering
\fontsize{10}{12}\selectfont
\sisetup{table-format = 3.2}
\resizebox{0.32\textwidth}{!}{%
\begin{tabular}{@{}cccc@{}}
\toprule
\textsc{AUG} & \textsc{BLEU} & \textsc{ROUGE-L} & \textsc{BLEURT} \\ \midrule
\xmark & \num{14.3} & \num{36.9} & \num{48.9} \\
\cmark & $\mathbf{14.7}$ & $\mathbf{37.2}$ & $\mathbf{49.0}$ \\ \bottomrule
\end{tabular}%
}
\caption{How2Sign test performance of \method~$_{800}^{\text{YT} + \text{H2S}}$ with and without finetuning augmentation.}
\label{tab:augmentation}
\end{table}

\paragraph{Should we augment data at finetuning?}
Augmentation such as flipping, cropping, and RandAug may improve generalization, but it comes at a high storage cost at finetuning time, as the video features are extracted offline. Is it worth the cost? We compared using 60 epochs of augmented videos (a 60-fold increase in storage) with not using any augmentation. The results in Table~\ref{tab:augmentation} show that using augmentation yields a reasonable \num{0.4} BLEU gain, suggesting that augmention can be useful when storage is not a major concern.

\begin{table}[ht!]
\centering
\fontsize{10}{12}\selectfont
\sisetup{table-format = 3.2}
\resizebox{0.48\textwidth}{!}{%
\begin{tabular}{@{}lccccc@{}}
\toprule
\textsc{Initialization} & \textsc{MAE} & {CLIP} & \textsc{BLEU} & \textsc{ROUGE-L} & \textsc{BLEURT} \\ \midrule
Hiera $_{800}^{\text{K400}}$ & \xmark & \cmark & \num{2.1} & \num{14.9} & \num{35.0} \\ \midrule
\multirow{3}{*}{\method $_{600}^{\text{YT} + \text{H2S}}$} & \xmark & \cmark & \num{11.0} & \num{32.1} & \num{44.7} \\
& \cmark & \xmark & \num{14.3} & \num{36.9} & \num{48.9} \\
& \cmark & \cmark & $\mathbf{15.5}$ & $\mathbf{38.4}$ & $\mathbf{49.6}$ \\ \bottomrule
\end{tabular}%
}
\caption{How2Sign test performance when including (\cmark) or removing (\xmark) the MAE and CLIP objectives and pretraining from the original Hiera $_{800}^{\text{K400}}$ or \method~$_{600}^{\text{YT} + \text{H2S}}$ checkpoint for 200 epochs on YT+H2S, followed by finetuning on the same data.}
\label{tab:loss_component_ablation}
\end{table}

\paragraph{Are both pretraining objectives necessary?}
In \S\ref{sec:main_results}, we saw that language-supervised video-text pretraining is highly effective when combined with MAE. Are both necessary? We compared pretraining for \num{200} epochs with either and both objectives, initializing from a \num{600}-epoch \method checkpoint, and also performed \num{200} epochs of CLIP-only pretraining from the original pretrained Hiera. The results in Table~\ref{tab:loss_component_ablation} show that removing either objective results in a performance drop. The drop is larger when removing MAE, indicating its continued importance after \num{600} epochs of MAE-only training. Initializing from the original Hiera results in very poor performance (\num{2.1} BLEU), suggesting that language-supervised pretraining alone is not useful in our setting. Considering that language supervision has previously been shown to be effective in isolation \citep{zhou-etal-2023-iccv}, this may be due to the FLIP-style masking and the fact that we do not jointly pretrain T5.
We also emphasize that language-supervised pretraining falls under stage II of our framework as it requires annotations; it can, therefore, only serve as an addition to self-supervised pretraining, but not a replacement.

\section{Conclusion}
Through controlled experiments, we studied the effectiveness of self-supervised pretraining for SLT while considering privacy risks. We introduce \method, a novel, scalable, and privacy-aware SLT method that leverages masked autoencoding on anonymized video. It achieves state-of-the-art ASL-to-English translation performance on the How2Sign benchmark, outperforming the best previous models in the finetuned and zero-shot setting by over \num{3} BLEU. 

Our results demonstrate the promise of self-supervised learning to alleviate the data scarcity issue and further scale up sign language processing in the future. We found that video anonymization, even via simple techniques such as face blurring, has relatively little negative impact on downstream performance, further proving that we can build more proficient systems without neglecting important privacy concerns. We hope that this work, alongside the code and data we release, will spur future developments that benefit the d/Deaf and hard of hearing communities.

\section*{Limitations}

\paragraph{Compute Requirements}

Currently, \method requires access to substantial compute to train at the scale of Youtube-ASL (\num{600}K videos). This is primarily due to the high dimensionality of video data, exacerbated by the long clip length and information density in sign language content, which creates a data-loading bottleneck and increases the memory footprint, especially in combination with a transformer architecture.
Our longest pretraining run in full precision (fp32) took approximately two weeks on \num{64} A\num{100} GPUs. We believe that it will be important to drive down this cost in the future and make large-scale video pretraining more accessible. While many simple interventions such as mixed precision, gradient accumulation, and gradient checkpointing could drastically reduce the memory footprint, they usually come at the cost of training time or stability.
In general, we note that this limitation is not unique to our approach but often not apparent due to training being conducted on nearly \num{100}$\times$ smaller datasets such as RWTH-Phoenix-Weather \num{2014}T \citep[\num{7}K videos; ][]{camgoz-etal-2018-neural}.

\paragraph{Anonymization}

We rely on face blurring for video anonymization, which is known to incur a loss of linguistic information (see \S\ref{sec:sign_anonymization}). In the future, it will be worth investigating more sophisticated methods, such as using synthetic appearances.
Also, largely due to a lack of linguistic tools for continuous signing, we did not investigate what effects anonymization may have on the translations \emph{qualitatively}. For instance, it would be instructive to know whether the model successfully disambiguates certain phonemes with similar manual features through context in the absence of facial information.

\paragraph{Languages}

Due to the availability of sufficiently large datasets for our pretraining experiments, we only experiment with American Sign Language and English, the two highest-resource signed and spoken languages. We aim to diversify this language selection in the future.

\section*{Ethics Statement}

Regarding performance, our models may contain demographic biases and underperform for certain race, gender, and age groups. For instance, even though the YouTube-ASL dataset (a dataset we used for pretraining and supervised finetuning) features over 2500 signers, the authors did not provide demographic details of these signers. Similarly, our DailyMoth-70h dataset includes only one signer (white, male, and early middle-aged). As such, it is unclear how our models perform for underrepresented users, who, aside from having diverse identities, may introduce different accents or regional variations of ASL that our models do not adequately capture. We call for future research on SLT to be more explicit about documenting demographic biases in their datasets and models.

Lastly, we emphasize that anonymization inherently does not offer any formal privacy guarantees in contrast to frameworks such as differential privacy \citep{dwork-etal-2006-dp}, which fundamentally comes at a (often substantial) cost in utility \citep{geng-etal-2020-analysis}. As such, while our work (and the use of facial obfuscation in general) represents an important first step towards comprehensively protecting the privacy of signers, it should not be relied on in sensitive or high-stakes applications.

\section*{Acknowledgements}

We thank members of the Seamless Communication team at FAIR for helpful feedback and discussions throughout this project. We also thank Florian Metze and Amanda Duarte for their prompt assistance with our questions about the How2Sign dataset.

\bibliography{custom}
\bibliographystyle{acl_natbib}

\appendix

\section{Pose landmarks vs RGB Video}
\label{app:pose_vs_video}

Pose landmarks (e.g., from MediaPipe Holistic) are cited as a privacy-preserving alternative to RGB video for SLT \citep{moryossef-etal-2021-immediate, uthus-etal-2023-youtube}. While they may indeed offer benefits in terms of efficiency and generalization, we argue that they do not offer meaningful privacy protection either. For instance, using a sufficiently large number of facial landmarks that are estimated accurately results in what is essentially a scan of the facial geometry, a biometric identifier according to legislation like the Biometric Information Privacy Act (BIPA).\footnote{\url{https://www.ilga.gov/legislation/ilcs/ilcs3.asp?ActID=3004&ChapterID=57}}
Despite abstracting away shallow information about a person's appearance, pose information could, therefore, be (mis)used in similar ways as de-anonymized video. Analogous to facial obfuscation in video, one could reduce the number of facial landmarks or add noise to them to hinder re-identification, but doing so also results in (arguably even more) loss of information.

\begin{table*}[t]
\centering
\resizebox{\textwidth}{!}{%
\begin{tabular}{@{}lllll@{}}
\toprule
\multicolumn{1}{c}{} & \textsc{Train} & \textsc{Validation} & \textsc{Test} & \textsc{Full} \\ \midrule
\addlinespace[0.3em]
\rowcol\multicolumn{5}{l}{\textbf{Raw data}} \\
\addlinespace[0.2em]
Number of signers & --- & --- & --- & \num{1} \\
Number of videos & --- & --- & --- & \num{496} \\
Video duration (hours) & --- & --- & --- & \num{76.9} \\
Time frame & 01/21--04/23 & 02/22--04/23 & 02/22--04/23 & 01/21--04/23 \\
\addlinespace[0.3em]
\rowcol\multicolumn{5}{l}{\textbf{Segmented data}} \\
\addlinespace[0.2em]
Number of clips & \num{41412} & \num{2789} & \num{4185} & \num{48386} \\
Clip duration (hours) & \num{65.8} & \num{4.0} & \num{6.0} & \num{75.8} \\
Vocabulary (words) & \num{18495} & \num{4803} & \num{6040} & \num{19694} \\
Duration in seconds per clip (mean / std / 90\textsuperscript{th} percentile) & \num{5.7} / \num{2.4} / \num{8.9} & \num{5.2} / \num{2.1} / \num{7.9} & \num{5.2} / \num{2.1} / \num{7.9} & \num{5.6} / \num{2.3} / \num{8.7} \\
Characters per caption (mean / std / 90\textsuperscript{th}  percentile) & \num{43.9} / \num{12.7} / \num{58.0} & \num{44.1} / \num{12.8} / \num{59.0} & \num{43.3} / \num{12.9} / \num{59.0} & \num{43.9} / \num{12.7} / \num{59.0} \\
Words per caption (mean / std / 90\textsuperscript{th}  percentile) & \num{8.6} / \num{2.4} / \num{12.0} & \num{8.7} / \num{2.4} / \num{12.0} & \num{8.5} / \num{2.4} / \num{12.0} & \num{8.6} / \num{2.4} / \num{12.0} \\ \bottomrule
\end{tabular}%
}
\caption{DailyMoth-70h dataset statistics}
\label{tab:dailymoth_statistics}
\end{table*}

\section{DailyMoth-70h Dataset}
\label{app:dailymoth_data}

We introduce \emph{DailyMoth-70h}, a dataset containing over \num{70}h of video with aligned English captions of a single native ASL signer (white, male, and early middle-aged) from the ASL news page TheDailyMoth.\footnote{\url{https://www.dailymoth.com/}} We obtained the data via a license agreement with the host of TheDailyMoth.

\paragraph{Download and License}
The fully self-contained dataset is available under a CC BY-NC license at \url{https://github.com/facebookresearch/ssvp_slt}.

\begin{figure*}[ht!]
\centering
\begin{subfigure}[b]{0.6\textwidth}
\includegraphics[width=\textwidth]{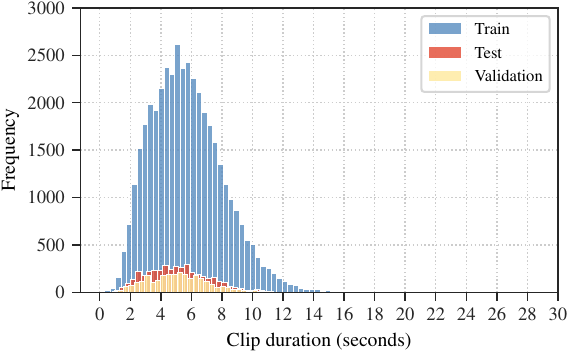}
\end{subfigure}

\vfill

\begin{subfigure}[b]{0.6\textwidth}
\includegraphics[width=\textwidth]{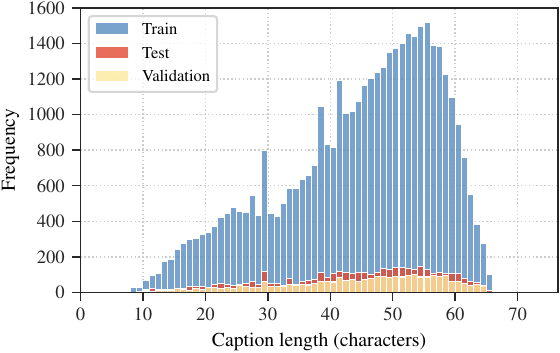}
\end{subfigure}

\vspace{0.3em}

\begin{subfigure}[b]{0.6\textwidth}
\includegraphics[width=\textwidth]{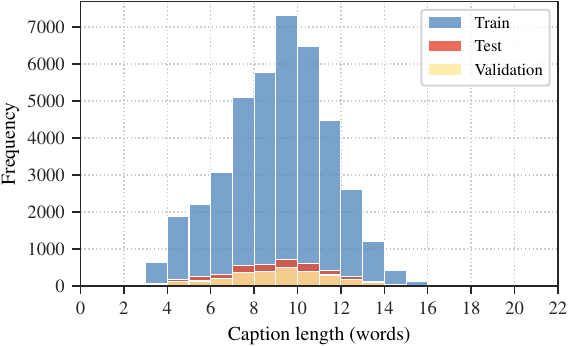}
\end{subfigure}

\caption{DailyMoth-70h dataset split statistics}
\label{fig:dailymoth_histograms}
\end{figure*}

\paragraph{Statistics}
We provide detailed dataset statistics in Table~\ref{tab:dailymoth_statistics} and Figure~\ref{fig:dailymoth_histograms}.

\paragraph{Purpose}
The dataset is fully self-contained and can therefore serve as a new benchmark for studying the task of single-signer translation (e.g., for building personalized systems). Furthermore, sign language translation involves overcoming several challenges such as generalizing over signers, their appearances and their signing styles as well as mapping spatio-temporal signed utterances to spoken words. DailyMoth-70h can be used to disentangle some of these challenges by eliminating the signer and style variances and allow researchers to ablate their models more focused on the sign-to-spoken language mapping.

\paragraph{Preprocessing}

We first segmented the raw videos into clips of \circa 5.6 seconds on average based on their aligned English captions. Each entry in the SubRip subtitle (SRT) file, which maps video timestamps to captions, was chosen to be a distinct datapoint. Accordingly, example clips are often sentence fragments rather than full sentences.

Afterwards, the segmented video clips were automatically cropped such that the signer is approximately in the center of the frame and resized to $224 \times 224$ pixels. The preprocessed clips were saved in their native framerates, which are either 24 or 30 fps.

Next, many videos had burned-in captions which, if not removed, would reduce the translation task to a simple OCR task. We, therefore, used an off-the-shelf text detection model to identify burned-in captions in the videos, and blurred the captions conservatively. Although the blurring may be imperfect due to errors made by the text detector, this intervention should nevertheless solve the concern of models shortcutting the SLT task for the most part.

Finally, we split the data into training, validation, and test splits. The proportions (\num{85}\% / \num{6}\% / \num{9}\%) were chosen to approximately match How2Sign. The validation and test examples were randomly sampled from the subset of videos posted after January 2022, which avoids data leakage from Youtube-ASL or OpenASL \citep{shi-etal-2022-open}, both of which have cut-off dates before/during January 2022, into the DailyMoth-70h evaluation splits. The training examples were randomly sampled from the full range of dates (January 2021 to April 2023).

\section{Reproducibility}

\subsection{Model and Training Configurations}
\label{app:training_details}

We report our pretraining configurations for \method in Table~\ref{tab:pretraining_settings} and \method-LSP in Table~\ref{tab:pretraining_settings_lsp}. Our finetuning configurations are listed  in Table~\ref{tab:finetuning_settings_yt} for Youtube-ASL (+ How2Sign) and Table~\ref{tab:finetuning_settings_h2s_dm} for How2Sign-only and DailyMoth-70h.

\subsection{Code}
Our implementation uses Python 3.10 and PyTorch 2.0.1 \citep{paszke-etal-2019-pytorch} compiled with CUDA 11.7. The code is available under a CC BY-NC license at \url{https://github.com/facebookresearch/ssvp_slt}.

\subsection{Hardware \& Runtime}
We ran our experiments on NVIDIA A100 \num{80}GB and V100 \num{32}GB GPUs. On Youtube-ASL (+ How2Sign), pretraining took \circa 20 minutes (\method) / 30 minutes (\method-LSP) per effective epoch on 64 A100 or 128 V100 GPUs. On How2sign or DailyMoth-70h, an effective epoch of \method pretraining took \circa 3 minutes.
Finetuning and evaluating T5 and BART on Youtube-ASL (+ How2Sign) took \circa 8 and 4 minutes, respectively, per epoch on 32 V100 GPUs. Training T5 was slower due to training in full precision as opposed to fp16 and using a smaller batch size with gradient accumulation. Finetuning and evaluating with \citet{tarres-etal-2023-instructional}'s setup on How2Sign or DailyMoth-70h took \circa 1 minute per epoch on a single V100 GPU.

\section{Qualitative Examples}

In Table~\ref{tab:qualitative_examples}, we provide qualitative examples of our best-performing model (15.5 BLEU on How2Sign), compared against the best-performing models from \citet{tarres-etal-2023-instructional}, \citet{uthus-etal-2023-youtube}, as well as the reference translations. The examples were picked from the How2Sign test split by \citet{tarres-etal-2023-instructional}. Examples (3)--(5) are, anecdotally, more challenging than the average test example. We find that our model is mostly on-topic and matches the syntactic structure, although it can still struggle with repetitions and the mixing-up of signs. Our model's failure patterns are more similar to \citet{uthus-etal-2023-youtube}'s models---possibly a result of finetuning the same base model (T5-v1.1) on the same datasets (Youtube-ASL and How2Sign). For instance, in example (3), both models mistakenly predict the verb ``feed'' (and mispredict everything that comes after) but flawlessly match the syntactic structure of the reference translation. Overall, both baselines appear to exhibit a higher occurrence of (complete) mis-translation, which aligns with our quantitative results.

\vfill

\begin{table}[ht!]
\centering
\fontsize{10}{12}\selectfont
\sisetup{table-format = 3.2}
\resizebox{0.4\textwidth}{!}{%
\begin{tabular}{@{}ll@{}}
\toprule
\textsc{parameter}            & \textsc{value} \\ \midrule
Decoder blocks & \num{8} \\
Decoder heads & \num{8} \\
Mask ratio & \num{0.9} \\
Drop path rate & \num{0.2} \\
Video size ($T$, $C$, $H$, $W$) & (\num{128}, \num{3}, \num{224}, \num{224}) \\
Sampling Rate & \num{2} \\
Face Blur & \cmark \\
Random Crop & \cmark \\
Horizontal Flip & \cmark \\
RandAug & \cmark (\num{4}, \num{7}) \\
Repeated sampling & \num{2} \\
Optimizer              & AdamW            \\
Optimizer momentum     & $\beta_1$ = \num{0.9}, $\beta_2$ = \num{0.95} \\
Weight decay               & \num{0.05}          \\
Peak learning rate         & \num{8e-4}   \\
Learning rate schedule & Cosine Decay                        \\
Minimum learning rate      & \num{1e-5}     \\
Effective warmup epochs & \num{120}            \\
Effective epochs             & \num{800}             \\
Effective batch size                 & \num{256}       \\
Gradient clipping & --- \\
Precision & fp32 \\
\bottomrule
\end{tabular}%
}
\caption{\textbf{\method} pretraining settings}
\label{tab:pretraining_settings}
\end{table}

\begin{table}[ht!]
\centering
\fontsize{10}{12}\selectfont
\sisetup{table-format = 3.2}
\resizebox{0.4\textwidth}{!}{%
\begin{tabular}{@{}ll@{}}
\toprule
\textsc{parameter}            & \textsc{value} \\ \midrule
CLIP Text tower & T5-v1.1-base \\
CLIP Feature pooling & mean \\
CLIP Feature projection & 2-layer MLP \\
Decoder blocks & \num{8} \\
Decoder heads & \num{8} \\
Mask ratio & \num{0.9} \\
Drop path rate & \num{0.2} \\
Video size ($T$, $C$, $H$, $W$) & (\num{128}, \num{3}, \num{224}, \num{224}) \\
Sampling Rate & \num{2} \\
Face Blur & \cmark \\
Random Crop & \cmark \\
Horizontal Flip & \cmark \\
RandAug & \cmark (\num{4}, \num{7}) \\
Repeated sampling & \num{1} \\
Optimizer              & AdamW            \\
Optimizer momentum     & $\beta_1$ = \num{0.9}, $\beta_2$ = \num{0.95} \\
Weight decay               & \num{0.05}          \\
GradNorm $\alpha$ & 1.0 \\
Peak learning rate         & M = \num{1e-4}, GN = \num{1e-2}   \\
Learning rate schedule & Cosine Decay                        \\
Minimum learning rate      & M = \num{1e-6}, GN = \num{1e-4}     \\
Effective warmup epochs & M = \num{10}, GN = \num{1}            \\
Effective epochs             & \num{200}             \\
Effective batch size                 & \num{512}       \\
Gradient clipping & 1.0 \\
Precision & fp32 \\

\bottomrule
\end{tabular}%
}
\caption{\textbf{\method-LSP} pretraining settings. ``M'' refers to the main optimizer while ``GN'' refers to the GradNorm optimizer.}
\label{tab:pretraining_settings_lsp}
\end{table}

\begin{table}[ht!]
\centering
\fontsize{10}{12}\selectfont
\sisetup{table-format = 3.2}
\resizebox{0.4\textwidth}{!}{%
\begin{tabular}{@{}ll@{}}
\toprule
\textsc{parameter}            & \textsc{value} \\ \midrule
Model \& Tokenizer & T5-v1.1 \\
Dropout probability & \num{0.3} \\
Label smoothing & \num{0.2} \\
Number of beams & \num{5} \\
Video size ($T$, $C$, $H$, $W$) & ($T$, \num{3}, \num{224}, \num{224}) \\
Sampling Rate & \num{2} \\
Face Blur & \cmark \\
Random Crop & \xmark \\
Horizontal Flip & \xmark \\
RandAug & \xmark \\
Min sequence length & \num{0} \\
Max sequence length & \num{1024} \\
Max target length & \num{128} \\
Optimizer              & AdamW            \\
Optimizer momentum     & $\beta_1$ = \num{0.9}, $\beta_2$ = \num{0.95} \\
Weight decay               & \num{1e-1}          \\
Peak learning rate         & \num{5e-4}   \\
Learning rate schedule & Cosine Decay                        \\
Minimum learning rate      & \num{1e-7}     \\
Warmup epochs & \num{2}            \\
Epochs             & \num{100}             \\
Batch size                 & \num{256}       \\
Early stopping & \cmark \\
Gradient clipping & \num{1.0} \\
Precision & fp32 \\

\bottomrule
\end{tabular}%
}
\caption{Finetuning settings for Youtube-ASL.}
\label{tab:finetuning_settings_yt}
\end{table}

\begin{table*}[ht!]
\centering
\fontsize{10}{12}\selectfont
\sisetup{table-format = 3.2}
\resizebox{0.4\textwidth}{!}{%
\begin{tabular}{@{}ll@{}}
\toprule
\textsc{parameter}            & \textsc{value} \\ \midrule
Encoder layers & \num{6} \\
Decoder layers & \num{3} \\
Attention heads & \num{4} \\
Embedding dim & \num{256} \\
FFN embedding dim & \num{1024} \\
Output dim & \num{256} \\
Layerdrop  & \num{0.0} \\
Activation function & ReLU \\
LayerNorm Before & \cmark \\
LayerNorm Embedding & \cmark \\
Scale embeddings & \cmark \\
Decoder share embeddings & \cmark \\
Vocab size & \num{7000} \\
Lowercase tokenizer & H2S = \cmark, DM = \xmark \\
Truecase outputs & H2S = \cmark, DM = \xmark \\
Dropout probability & \num{0.3} \\
Label smoothing & \num{0.2} \\
Number of beams & \num{5} \\
Video size ($T$, $C$, $H$, $W$) & ($T$, \num{3}, \num{224}, \num{224}) \\
Sampling Rate & \num{2} \\
Face Blur & H2S = \cmark, DM = \cmark / \xmark \\
Random Crop & \xmark \\
Horizontal Flip & \xmark \\
RandAug & \xmark \\
Min sequence length & \num{0} \\
Max sequence length & \num{1024} \\
Max target length & \num{128} \\
Optimizer              & AdamW            \\
Optimizer momentum     & $\beta_1$ = \num{0.9}, $\beta_2$ = \num{0.95} \\
Weight decay               & \num{1e-1}          \\
Peak learning rate         & \num{1e-2}   \\
Learning rate schedule & Cosine Decay                        \\
Minimum learning rate      & \num{1e-4}     \\
Warmup epochs & \num{10}            \\
Epochs             & \num{200}             \\
Batch size                 & \num{32}       \\
Early stopping & \cmark \\
Gradient clipping & \num{1.0} \\
Precision & fp16 \\

\bottomrule
\end{tabular}%
}
\caption{Finetuning settings for How2Sign (H2S) \& DailyMoth-70h (DM).}
\label{tab:finetuning_settings_h2s_dm}
\end{table*}

\begin{figure*}[ht]
    \centering
    \begin{subfigure}[b]{\textwidth}
        \centering
        \begin{minipage}{0.16\textwidth}
            \centering
            \includegraphics[width=\linewidth]{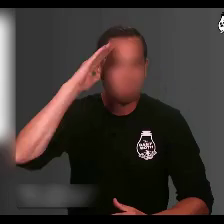}
        \end{minipage}\hfill
        \begin{minipage}{0.16\textwidth}
            \centering
            \includegraphics[width=\linewidth]{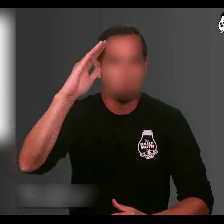}
        \end{minipage}\hfill
        \begin{minipage}{0.16\textwidth}
            \centering
            \includegraphics[width=\linewidth]{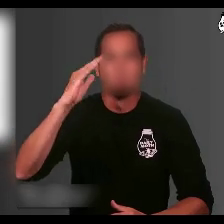}
        \end{minipage}\hfill
        \begin{minipage}{0.16\textwidth}
            \centering
            \includegraphics[width=\linewidth]{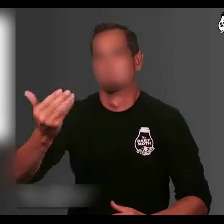}
        \end{minipage}\hfill
        \begin{minipage}{0.16\textwidth}
            \centering
            \includegraphics[width=\linewidth]{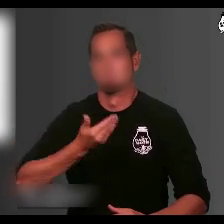}
        \end{minipage}\hfill
        \begin{minipage}{0.16\textwidth}
            \centering
            \includegraphics[width=\linewidth]{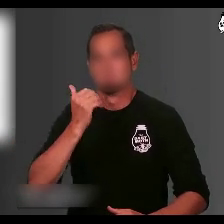}
        \end{minipage}

        \vspace{0.2cm}

        \begin{minipage}{0.16\textwidth}
            \centering
            \includegraphics[width=\linewidth]{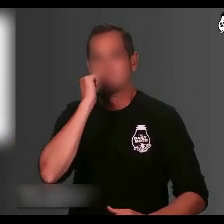}
        \end{minipage}\hfill
        \begin{minipage}{0.16\textwidth}
            \centering
            \includegraphics[width=\linewidth]{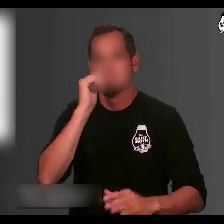}
        \end{minipage}\hfill
        \begin{minipage}{0.16\textwidth}
            \centering
            \includegraphics[width=\linewidth]{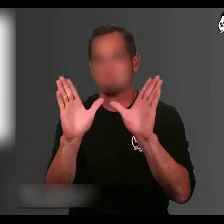}
        \end{minipage}\hfill
        \begin{minipage}{0.16\textwidth}
            \centering
            \includegraphics[width=\linewidth]{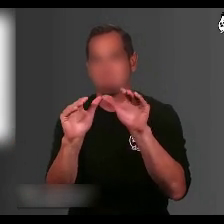}
        \end{minipage}\hfill
        \begin{minipage}{0.16\textwidth}
            \centering
            \includegraphics[width=\linewidth]{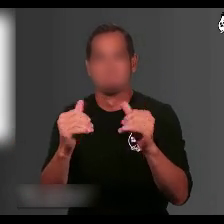}
        \end{minipage}\hfill
        \begin{minipage}{0.16\textwidth}
            \centering
            \includegraphics[width=\linewidth]{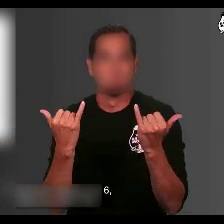}
        \end{minipage}
        \caption{English translation: ``\emph{Hello, welcome to The Daily Moth.}''}
    \end{subfigure}

    \vspace{1cm}

    \begin{subfigure}[b]{\textwidth}
        \centering
        \begin{minipage}{0.16\textwidth}
            \centering
            \includegraphics[width=\linewidth]{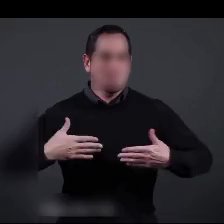}
        \end{minipage}\hfill
        \begin{minipage}{0.16\textwidth}
            \centering
            \includegraphics[width=\linewidth]{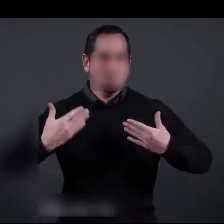}
        \end{minipage}\hfill
        \begin{minipage}{0.16\textwidth}
            \centering
            \includegraphics[width=\linewidth]{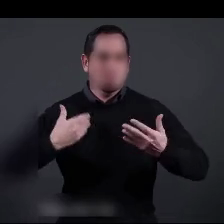}
        \end{minipage}\hfill
        \begin{minipage}{0.16\textwidth}
            \centering
            \includegraphics[width=\linewidth]{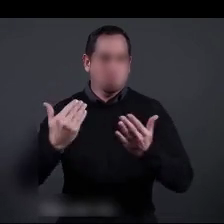}
        \end{minipage}\hfill
        \begin{minipage}{0.16\textwidth}
            \centering
            \includegraphics[width=\linewidth]{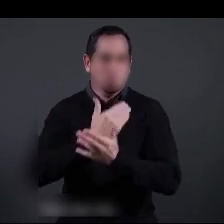}
        \end{minipage}\hfill
        \begin{minipage}{0.16\textwidth}
            \centering
            \includegraphics[width=\linewidth]{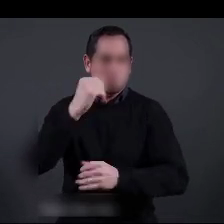}
        \end{minipage}

        \vspace{0.2cm}

        \begin{minipage}{0.16\textwidth}
            \centering
            \includegraphics[width=\linewidth]{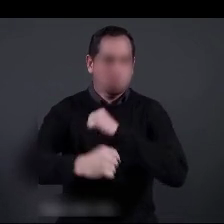}
        \end{minipage}\hfill
        \begin{minipage}{0.16\textwidth}
            \centering
            \includegraphics[width=\linewidth]{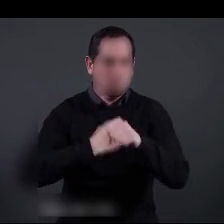}
        \end{minipage}\hfill
        \begin{minipage}{0.16\textwidth}
            \centering
            \includegraphics[width=\linewidth]{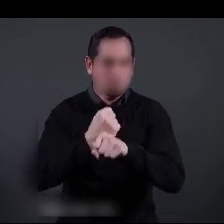}
        \end{minipage}\hfill
        \begin{minipage}{0.16\textwidth}
            \centering
            \includegraphics[width=\linewidth]{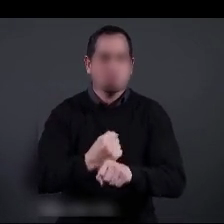}
        \end{minipage}\hfill
        \begin{minipage}{0.16\textwidth}
            \centering
            \includegraphics[width=\linewidth]{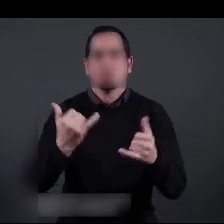}
        \end{minipage}\hfill
        \begin{minipage}{0.16\textwidth}
            \centering
            \includegraphics[width=\linewidth]{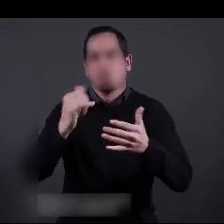}
        \end{minipage}
        \caption{English translation: ``\emph{Happy New Year}''}
    \end{subfigure}

    \caption{Example frames sampled from two videos in the blurred version of the DailyMoth-70h training split.}
    \label{fig:example_blurred_frames}
\end{figure*}

\begin{table*}[ht!]
\centering
\fontsize{10}{12}\selectfont
\resizebox{\textwidth}{!}{%
\begin{tabular}{@{}llp{12cm}@{}}
\toprule
\multirow{4}{*}{(1)} & \textbf{Reference} & And that's a great vital point technique for women's self defense. \\
& Tarr\'{e}s et al. & It's really a great point for women's self defense. \\
& Uthus et al. & It's really great for women's self defense. \\
& Ours & This is a really great point for women's self defense. \\ \midrule
\multirow{4}{*}{(2)} & \textbf{Reference} & In this clip I'm going to show you how to tape your cables down. \\
& Tarr\'{e}s et al. & In this clip I'm going to show you how to improve push ups. \\
& Uthus et al. & In this clip we're going to show you how to cut a piece of clay. \\
& Ours & In this clip I'm going to show you how to clip the cable, the cable. \\ \midrule
\multirow{4}{*}{(3)} & \textbf{Reference} & In this segment we're going to talk about how to load your still for distillation of lavender essential oil. \\
& Tarr\'{e}s et al. & Ok, in this clip, we're going to talk about how to fold the ink for the lid of the oil. \\
& Uthus et al. & In this clip we're going to talk about how to feed a set of baiting lizards for a lava field oil. \\
& Ours & In this clip we're going to talk about how to feed the trail for draining clean for laborer oil. \\ \midrule
\multirow{4}{*}{(4)} & \textbf{Reference} & You are dancing, and now you are going to need the veil and you are going to just grab the veil as far as possible. \\
& Tarr\'{e}s et al. & So, once you're belly dancing, once you've got to have the strap, you're going to need to grab the thumb, and try to avoid it. \\
& Uthus et al. & Their hopping and dancing is now, they're going to need their squat and squat and they're going to be able to move independently. \\
& Ours & So that she's going to get her hips up as far as she can, and now she's going to lift her head up as far as possible. \\ \midrule
\multirow{4}{*}{(5)} & \textbf{Reference} & But if you have to setup a new campfire, there’s two ways to do it in a very low impact; one is with a mound fire, which we should in the campfire segment earlier and the other way to setup a low impact campfire is to have a fire pan, which is just a steel pan like the top of a trash can. \\
& Tarr\'{e}s et al. & And other thing I'm going to talk to you is a little bit more space, a space that's what it's going to do, it's kind of a quick, and then I don't want to take a spray skirt off, and then I don't want it to take it to the top of it. \\
& Uthus et al. & But if you have to set up a new campfire, there are two ways to do a low impact fire, one is a cone fire, which we have to do in the tent earlier, and the other one is to set up a campfire in a fire pan. \\
& Ours & But if you have to set up a new campfire, this is one way to do it in a low impact. One is a monk fire. One is a campfire. The other one is to set a campfire in a campfire. That's just a post like the top of the post.
\\ \midrule
\multirow{4}{*}{(6)} & \textbf{Reference} & So, this is a very important part of the process. \\
& Tarr\'{e}s et al. & It's a very important part of the process. \\
& Uthus et al. & Alright, let's get started. \\
& Ours & It's an important part of the process. \\ \bottomrule
\end{tabular}%
}
\caption{Qualitative translation examples from our best-performing model compared to \citet{tarres-etal-2023-instructional}, \citet{uthus-etal-2023-youtube}, and the reference translations. The examples were picked from the How2Sign test set by \citet{tarres-etal-2023-instructional} and do not necessarily accurately reflect progress on the task. We see that our model is mostly on-topic, but can still struggle with repetitions and the mixing-up of signs.}
\label{tab:qualitative_examples}
\end{table*}
\end{document}